\documentclass[runningheads]{llncs}
\usepackage[T1]{fontenc}
\usepackage{graphicx,verbatim}

\usepackage{booktabs}
\usepackage{float}
\usepackage{xcolor}
\usepackage{enumitem}
\usepackage{amsmath}
\usepackage{amssymb}
\usepackage{bbm}
\usepackage{multirow}
\usepackage{tabularx}
\usepackage[table]{xcolor}
\begin{document}

%%%%%%%%%%%%%%%%%%%%%%%%%%%%%%%%%%%%%%%%%%%%%%%%
% Previous
% \title{CoRe-BT: Joint Radiology-Pathology Learning for Multimodal Brain Tumor Typing }
% \titlerunning{Joint Radiology-Pathology Learning for Multimodal Brain Tumor Typing}
% New
\title{CoRe-BT: A Multimodal Radiology-Pathology-Text Benchmark for Robust Brain Tumor Typing}
\titlerunning{CoRe-BT: A Multimodal Benchmark for Robust Brain Tumor Typing }
%% OR 
%% CoRe-BT: A Cross-Modal Radiology-Pathology Benchmark for Multimodal Brain Tumor Typing

%%%%%%%%%%%%%%%%%%%%%%%%%%%%%%%%%%%%%%%%%%%%%%%%
\makeatletter
\newcommand{\printfnsymbol}[1]{%
  \textsuperscript{\@fnsymbol{#1}}%
}
\makeatother

% \begin{comment}  %% Removed for anonymized MICCAI submission
% \author{Juampablo E. Heras Rivera\inst{1}\orcidID{0000-0002-0205-6329} \and
% Daniel K. Low\inst{1}\orcidID{0000-0002-9519-8691} \and
% Xavier Xiong \inst{1}\and
% Jacob J. Ruzevick\inst{1}\and
% Daniel D. Child\inst{1}\and
% Wen-wai Yim\inst{2}\orcidID{0000-0001-9011-0817}\and
% Mehmet Kurt\inst{1}\orcidID{0000-0002-5618-0296}\and
% Asma Ben Abacha\inst{2}\orcidID{0000-0001-6312-9387}}
\author{
Juampablo E. Heras Rivera\inst{1}\orcidID{0000-0002-0205-6329}\thanks{Equal contribution} \and
Daniel K. Low\inst{1}\orcidID{0000-0002-9519-8691}\printfnsymbol{1} \and
Xavier Xiong \inst{1} \and
Jacob J. Ruzevick\inst{1} \and
Daniel D. Child\inst{1} \and
Wen-wai Yim\inst{2}\orcidID{0000-0001-9011-0817} \and
Mehmet Kurt\inst{1}\orcidID{0000-0002-5618-0296}\thanks{Shared senior authorship} \and
Asma Ben Abacha\inst{2}\orcidID{0000-0001-6312-9387}\printfnsymbol{2}
}
%
% \Name{Jacob Ruzevick\nametag{$^{2}$}} \Email{ruzevick@uw.edu}\\
\authorrunning{J.E. Heras Rivera, D.K. Low et al.}
% First names are abbreviated in the running head.
% If there are more than two authors, 'et al.' is used.
%
\institute{University of Washington, Seattle, Washington 98105, USA \and
University of Washington School of Medicine, Seattle, Washington 98105, USA \and
Microsoft Health AI, Redmond, Washington 98052-6399, USA\\ }

\author{
Juampablo E. Heras Rivera\inst{1}\orcidID{0000-0002-0205-6329}\thanks{Equal contribution} \and
Daniel K. Low\inst{2}\orcidID{0000-0002-9519-8691}\printfnsymbol{1} \and
Xavier Xiong \inst{2} \and
Jacob J. Ruzevick\inst{2} \and
Daniel D. Child\inst{2} \and
Wen-wai Yim\inst{3}\orcidID{0000-0001-9011-0817} \and
Mehmet Kurt\inst{1}\orcidID{0000-0002-5618-0296}\thanks{Shared senior authorship} \and
Asma Ben Abacha\inst{3}\orcidID{0000-0001-6312-9387}\printfnsymbol{2}
}

% \and
% ABC Institute, Rupert-Karls-University Heidelberg, Heidelberg, Germany\\
% \email{\{abc,lncs\}@uni-heidelberg.de}}

% \end{comment}

%% Daniel K Low, Juampablo E. Heras Rivera, Wen-wai Yim, Jacob Ruzevick, Xavier Xiong, Mehmet Kurt*, Asma Ben Abacha*  

%% * Shared last authorship

% \author{Anonymized Authors}  %% Added for anonymized MICCAI submission
% \authorrunning{Anonymized Author et al.}
% \institute{Anonymized Affiliations \\
%     \email{email@anonymized.com}}

%%%%%%%%%%%%%%%%%%%%%%%%%%%%%%%%%%%%%%%%%%%%%%%%
  
\maketitle             
\begin{abstract}
Accurate brain tumor typing requires integrating heterogeneous clinical evidence, including magnetic resonance imaging (MRI), histopathology, and pathology reports, which are often incomplete at the time of diagnosis. We introduce CoRe-BT, a cross-modal radiology-pathology-text benchmark for brain tumor typing, designed to study robust multimodal learning under missing modality conditions. The dataset comprises 310 patients with multi-sequence brain MRI (T1, T1c, T2, FLAIR), including 95 cases with paired H\&E-stained whole-slide pathology images and pathology reports. All cases are annotated with tumor type and grade, and MRI volumes include expert-annotated tumor masks, enabling both region-aware modeling and auxiliary learning tasks. Tumors are categorized into six clinically relevant classes capturing the heterogeneity of common and rare glioma subtypes. We evaluate tumor typing under variable modality availability by comparing MRI-only models with multimodal approaches that incorporate pathology information when present. Baseline experiments demonstrate the feasibility of multimodal fusion and highlight complementary modality contributions across clinically relevant typing tasks. CoRe-BT provides a grounded testbed for advancing multimodal glioma typing and representation learning in realistic scenarios with incomplete clinical data.  


\end{abstract}

\section{Introduction}

Brain tumor typing is a central task in neuro-oncology, directly informing treatment selection, prognostic assessment, and clinical trial eligibility. Diagnostic workflows integrate multiple sources of evidence, including magnetic resonance imaging (MRI), histopathologic examination of hematoxylin and eosin (H\&E)-stained tissue sections, and narrative diagnostic reports. These modalities capture complementary aspects of tumor biology across scales, from macroscopic radiologic features to microscopic histomorphologic patterns and higher-level diagnostic interpretation.

Recent progress in foundation models and vision-language models (VLMs) has enabled large-scale multimodal representation learning across imaging and text. Contrastive pretraining frameworks such as CLIP \cite{RadfordKHRGASAM21} and medical adaptations including BioMedCLIP \cite{zhang2025biomedclip} have demonstrated the value of aligned visual-text embeddings in medical domains. More recently, multimodal foundation models such as Florence \cite{yuan2021florence} and domain-adapted systems including MedImageInsight \cite{medimageinsight2024} have advanced cross-modal reasoning across heterogeneous visual inputs and clinical language.

In computational pathology, hierarchical and weakly supervised whole-slide modeling approaches, including the Hierarchical Image Pyramid Transformer (HIPT) \cite{chen2022hipt} and foundation-scale pathology encoders such as GigaPath \cite{xu2024gigapath}, have demonstrated strong performance for tumor classification from gigapixel whole-slide images. Despite these advances, most prior studies focus on unimodal settings, coarse diagnostic groupings, or assume complete modality availability at inference time. In routine clinical practice, however, imaging is typically available at presentation, whereas histopathologic evaluation and finalized diagnostic reporting may occur later in the care pathway.

Robust brain tumor typing therefore requires multimodal models that can integrate cross-scale information while remaining resilient to incomplete inputs. In this work, we focus on glioma, the most common primary malignant brain tumor in adults, whose diagnosis and subtyping rely on integrated radiologic and histopathologic assessment \cite{Louis-2021,Price-2024}. We present a clinically grounded multimodal benchmark, CoRe-BT\footnote{The benchmark will be made available upon request and completion of a data usage agreement, in compliance with institutional and hospital data governance policies.}, for glioma tumor typing that integrates MRI, whole-slide histopathology, and diagnostic text. The benchmark supports variable modality availability at inference time and employs a pathologist-validated hierarchical classification schema to enable fine-grained and clinically meaningful tumor categorization under realistic clinical conditions.

Our contributions are threefold:
(1) We introduce CoRe-BT, a clinically grounded multimodal benchmark for fine-grained glioma tumor typing that integrates radiology, histopathology, and diagnostic text within a unified framework.
(2) We formalize evaluation under variable modality availability, enabling systematic assessment of multimodal models in missing-modality settings that reflect real-world clinical workflows.
(3) We provide standardized baseline evaluations and modality ablation analyses within a pathologist-validated hierarchical labeling framework.

\section{Related Work}

% Datasets:
% \begin{itemize}
%     \item MICCAI challenge \cite{Farahani2020CPMRP_miccai20challenge} [x]
%     \item MICCAI challenge winners \cite{wang2022combining_challengewinners} 
%     \item TCGA dataset [x]
%     https://www.cancerimagingarchive.net/collection/tcga-gbm/ [X]
%     \item Foundation model approaches
%     - pathology only: \cite{saueressig2025histologyFM}
%     - mri only: \cite{li2025MRIfoundation}  
%     % \item Segmentation \cite{DBLP:conf/midl/RenSRRHCK24,DBLP:journals/corr/abs-2508-16004}... 
% \end{itemize}
% $\rightarrow$ No approaches with dual foundation models
%% Most relevant dataset: https://zenodo.org/records/3718894

Large-scale public MRI datasets have substantially advanced automated glioma analysis. The RSNA-ASNR-MICCAI BraTS challenge series \cite{baid2021rsna} established standardized multi-parametric MRI benchmarks for tumor segmentation and outcome prediction, becoming the primary evaluation platform for brain tumor imaging models. However, BraTS is restricted to radiology data and does not provide paired whole-slide histopathology or structured diagnostic text. Radiogenomic collections such as TCGA-GBM\footnote{\url{https://www.cancerimagingarchive.net/collection/tcga-gbm/}} and TCGA-LGG\footnote{\url{https://www.cancerimagingarchive.net/collection/tcga-lgg/}}, include imaging and molecular annotations, but they were not designed as standardized benchmarks for multimodal learning under controlled missing-modality conditions.

Recent efforts have explored multimodal glioma classification using paired radiology and histopathology data, notably in the context of the Computational Precision Medicine Radiology--Pathology challenge at MICCAI 2020 \cite{Farahani2020CPMRP_miccai20challenge}. Such initiatives demonstrated the feasibility of integrating MRI and whole-slide histopathology for automated tumor classification. However, approaches on these datasets typically focus on a limited set of broad diagnostic categories and assume complete modality availability at inference time. Moreover, most prior studies are developed within controlled challenge settings and do not evaluate robustness to missing modalities or variability in clinical data availability.

In contrast, neuro-oncology workflows frequently involve incomplete multimodal information, where imaging is available early while histopathologic evaluation and finalized reports may be delayed. Additionally, fine-grained, clinically validated hierarchical tumor classifications remain underexplored in multimodal learning settings. To address these gaps, we introduce a clinically grounded benchmark for fine-grained glioma tumor type prediction that integrates MRI, whole-slide histopathology, and diagnostic text, while explicitly supporting variable modality availability. By combining cross-scale integration, missing-modality robustness, and pathologist-validated hierarchical labels, this work establishes a new evaluation framework for multimodal glioma classification.

%====================================================
\section{Task Definition}
%====================================================

%%Tumor Typing Correction: This task focuses on predicting or correcting tumor type labels by jointly analyzing brain MRI scans (including T1, T1c, T2, and FLAIR), H\&E-stained pathology slides, and pathology reports. The objective is to develop models that can integrate visual and textual information from radiology and pathology to arrive at an accurate tumor classification. This is a challenging problem because it requires reasoning across multiple types of data that differ in resolution, format, and clinical context. A successful solution could improve diagnostic accuracy and reduce delays by enabling earlier decision-making while molecular biomarker results are still pending.

We introduce a clinically grounded multimodal benchmark for glioma tumor type prediction. The task evaluates a model’s ability to perform cross-modal reasoning across radiology, histopathology, and clinical text, while remaining robust to missing modalities at inference time.

%====================================================
\subsection{Problem Formulation}
%====================================================

Each patient case may contain up to three modalities:

% \begin{itemize}
%     \item \textbf{MRI}: Pre-operative brain MRI including T1, contrast-enhanced T1 (T1c), T2, and FLAIR sequences;
%     \item \textbf{WSI}: Hematoxylin and eosin (H\&E)-stained whole-slide histopathology images;
%     \item \textbf{Report}: A free-text pathology report authored by a neuropathologist.
% \end{itemize}
\begin{itemize}[topsep=2pt, partopsep=0pt, parsep=0pt, itemsep=4pt, leftmargin=1.2em]
    \item \textbf{MRI}: Pre-operative brain MRI including T1, contrast-enhanced T1 (T1c), T2, and FLAIR sequences;
    \item \textbf{WSI}: H\&E-stained whole-slide histopathology images;
    \item \textbf{Report}: A free-text pathology report authored by a neuropathologist.
\end{itemize}
Let $\mathcal{M} = \{m_{\text{MRI}}, m_{\text{WSI}}, m_{\text{TXT}}\}$ denote the set of possible modalities. 
For each patient $i$, a subset of modalities $\mathcal{S}_i \subseteq \mathcal{M}$ may be observed at inference time.

The model learns a classifier

\begin{equation}
f: \mathcal{X}_{\mathcal{S}_i} \rightarrow \mathcal{Y},
\end{equation}

where $\mathcal{X}_{\mathcal{S}_i}$ denotes the input representation constructed from the available modalities and $\mathcal{Y}$ represents the set of clinically derived tumor classes.

Importantly, a single unified model is trained to operate under variable modality availability, reflecting real-world clinical workflows in which histopathologic data may be delayed or unavailable at the time of initial imaging-based decision-making.

This task requires integrating signals across multiple representational scales:

\begin{itemize}
    \item Macroscopic radiologic features derived from 3D MRI volumes,
    \item Microscopic histomorphologic patterns observed in whole-slide histopathology images,
    \item Diagnostic interpretations expressed in free-text pathology reports.
\end{itemize}

These modalities differ substantially in spatial resolution, statistical structure, and semantic abstraction level, posing challenges for effective multimodal integration.

%====================================================
\subsection{Label Space and Hierarchical Organization}
%====================================================

Tumor labels were derived from integrated clinical, radiologic, and pathologic diagnoses and organized into a clinically meaningful two-level hierarchy. The final classification schema and label assignments were reviewed and validated by a board-certified neuropathologist to ensure clinical consistency and diagnostic accuracy. 

Table~\ref{tab:label_hierarchy} summarizes the hierarchical organization of tumor labels used in this study (with subtype counts). The classification reflects established neuropathologic diagnostic groupings and consolidates clinically related subtypes into unified categories for model training and evaluation. For the primary task, evaluation is conducted at the coarse-grained (Level 1) category level, while subtype information is preserved for future hierarchical modeling and analysis.

% \begin{table}[H]
% \centering
% \caption{Hierarchical tumor classification used for prediction. Evaluation is performed at the Level 1 category.}
% \footnotesize
% \setlength{\tabcolsep}{4pt}
% \renewcommand{\arraystretch}{1.05}

% \begin{tabular}{p{0.42\linewidth} p{0.44\linewidth} r}
% \toprule
% \textbf{Level 1 Class} & \textbf{Subtypes (count)} & \textbf{Total} \\
% \midrule

% \textbf{GBM / IDH-wt HGG} &
% GBM (153); recurrent (17); treatment-related (1); HGG NOS (1); rare (4)
% & 176 \\

% \textbf{IDH-mutant Astrocytoma} &
% Astrocytoma, IDH-mutant (71)
% & 71 \\

% \textbf{Oligodendroglioma} \textit{(IDH-mut., 1p/19q-codel.)} &
% Oligodendroglioma (30); Oligosarcoma (1)
% & 31 \\

% \textbf{Circumscribed Astrocytic Glioma} \textit{(MAPK-driven)} &
% Pilocytic (8); recurrent (3); BRAF-fusion (2)
% & 13 \\

% \textbf{Midline / H3-altered Glioma} &
% Diffuse midline (8); H3-3A mutant (1)
% & 9 \\

% \textbf{Glioneuronal / Neuronal Tumors} &
% SEGA (3); RGNT (2); Ganglioglioma (1)
% & 6 \\

% \textbf{Pediatric Diffuse Astrocytoma} &
% MYB/MYBL1-altered (1)
% & 1 \\

% \textbf{Other / NEC} &
% Low-grade glial NEC (1); NTRK-fusion (1); NF1-inactivated (1)
% & 3 \\

% \midrule
% \textbf{Total} & & \textbf{310} \\
% \bottomrule
% \end{tabular}

% \label{tab:label_hierarchy}
% \end{table}

\begin{table}[H]
\centering
\caption{Hierarchical tumor classification labels present in the CoRe-BT dataset. Evaluation is performed at the Level 1 category.}
\scriptsize
\setlength{\tabcolsep}{4pt}
\renewcommand{\arraystretch}{1.05}

\begin{tabular}{p{0.42\linewidth} p{0.44\linewidth} r}
\toprule
\textbf{Level 1 Class} & \textbf{Subtypes (count)} & \textbf{Total} \\
\midrule

\textbf{GBM / IDH-wt HGG} &
GBM (153); recurrent (17); treatment-related (1); HGG NOS (1); rare (4)
& 176 \\

\textbf{IDH-mutant Astrocytoma} &
Astrocytoma, IDH-mutant (71)
& 71 \\

\textbf{Oligodendroglioma} \textit{(1p/19q-codel., IDH-mut.)} &
Oligodendroglioma (30); Oligosarcoma (1)
& 31 \\

\textbf{Circumscribed Astrocytic Glioma} \textit{(MAPK-driven)} &
Pilocytic (8); recurrent (3); BRAF-fusion (2)
& 13 \\

\textbf{Midline / H3-altered Glioma} &
Diffuse midline (8); H3-3A mutant (1)
& 9 \\

\textbf{Glioneuronal / Neuronal Tumors} &
SEGA (3); RGNT (2); Ganglioglioma (1)
& 6 \\

\textbf{Pediatric Diffuse Astrocytoma} &
MYB/MYBL1-altered (1)
& 1 \\

\textbf{Other / NEC} &
Low-grade glial NEC (1); NTRK-fusion (1); NF1-inactivated (1)
& 3 \\

\midrule
\textbf{Total} & & \textbf{310} \\
\bottomrule
\end{tabular}

\label{tab:label_hierarchy}
\end{table}

%==============================
\section{Data Creation}
\subsection{MRI and Pathology Dataset}
Glioma patients from the University of Washington were selected over a two-year period (08/2023 - 08/2025), totaling 310 patients. Inclusion criteria required a confirmed histopathologic diagnosis of a glioma. Cases included mostly infiltrative gliomas, as well as CNS WHO grade 1 tumors,  primarily pilocytic astrocytomas. Pre-tumor resection imaging was obtained for each subject and organized into groups with full ($n=250$) or incomplete ($n=60$) MRI contrast sets (T1, T1c, T2, FLAIR), which are standard for glioma imaging and segmentation \cite{baid2021rsna}. Within the group with full MRI contrast sets, subjects were further divided into those with ($n=95$) or without ($n=155$) digital pathology imaging. This study was conducted under Institutional Review Board (IRB) approval.  %% UW Medicine hospitals

\subsection{Data preprocessing} 
\subsubsection{Histopathology} A total of 597 gigapixel WSIs are included in the dataset, corresponding to an average of over 5 slides per subject. To remove artifacts and annotations from WSIs, a custom HSV color segmentation method was applied which extracts tissue regions. Additionally, to reduce memory consumption and accelerate tiling, the CLAM \cite{CLAMlu2021data} library was used to process images.

\subsubsection{MRI} MRI images of all subjects were pre-processed using CaPTk to the standardized BraTS 2017-2023 pipeline \cite{Davatzikos2018CIPT,Pati2020CaPTk}. The steps were as follows: DICOM to NIfTI conversion, SRI24 co-registration, 1 mm isotropic resampling, and skull-stripping.

% \begin{table}[h]
% \centering
% \footnotesize
% \setlength{\tabcolsep}{4pt}
% \caption{Level 1 Grouped Distribution}
% \begin{tabularx}{\columnwidth}{@{}Xcccc@{}}
% \toprule
% \textbf{Description} & \textbf{Train} & \textbf{Val} & \textbf{Test} & \textbf{Total} \\ \midrule
% GBM / IDH-wt HGG & 99 & 20 & 18 & 137 \\ \addlinespace[2pt]
% IDH-mutant Diffuse Glioma & 67 & 11 & 9 & 87 \\ \addlinespace[2pt]
% Non-diffuse / Pediatric / Midline Glioma & 15 & 4 & 0 & 19 \\ \addlinespace[2pt]
% Other / NEC & 3 & 3 & 1 & 7 \\ \bottomrule
% \end{tabularx}
% \end{table}

\subsection{MRI Segmentation and Tumor Typing Ground Truth}
Ground truth labels for the three standard glioma subregions (edema, enhancing tumor, necrotic core) for subjects with full MRI contrast sets were created using state-of-the-art automated segmentation and expert annotator corrections. Segmentation masks were generated from the BraTS 2023-winning submission \cite{ferreira2024we}, which is based on an nnU-Net \cite{isensee2021nnunet} ensemble. Annotators with substantial glioma segmentation experience, following the BraTS region identification protocol \cite{baid2021rsna} reviewed each automated segmentation, making corrections to erroneously identified regions. The high-performing automated segmentation and input from annotators were combined to form the ground truth masks for glioma cases. Ground truth tumor typing for all glioma cases was derived from the final diagnosis and grade given by attending neuropathologists. %%% UW attending neuropathologists

%==============================

\section{Methods}

For multimodal brain tumor typing, we propose \textit{CoRe-BT-Fusion}, a dual foundation model framework. \textit{CoRe-BT-Fusion} integrates pretrained representations from large-scale histopathology and neuroimaging foundation models and performs supervised tumor classification via multimodal feature fusion.
\begin{figure}
    \centering
    \includegraphics[width=0.8\linewidth]{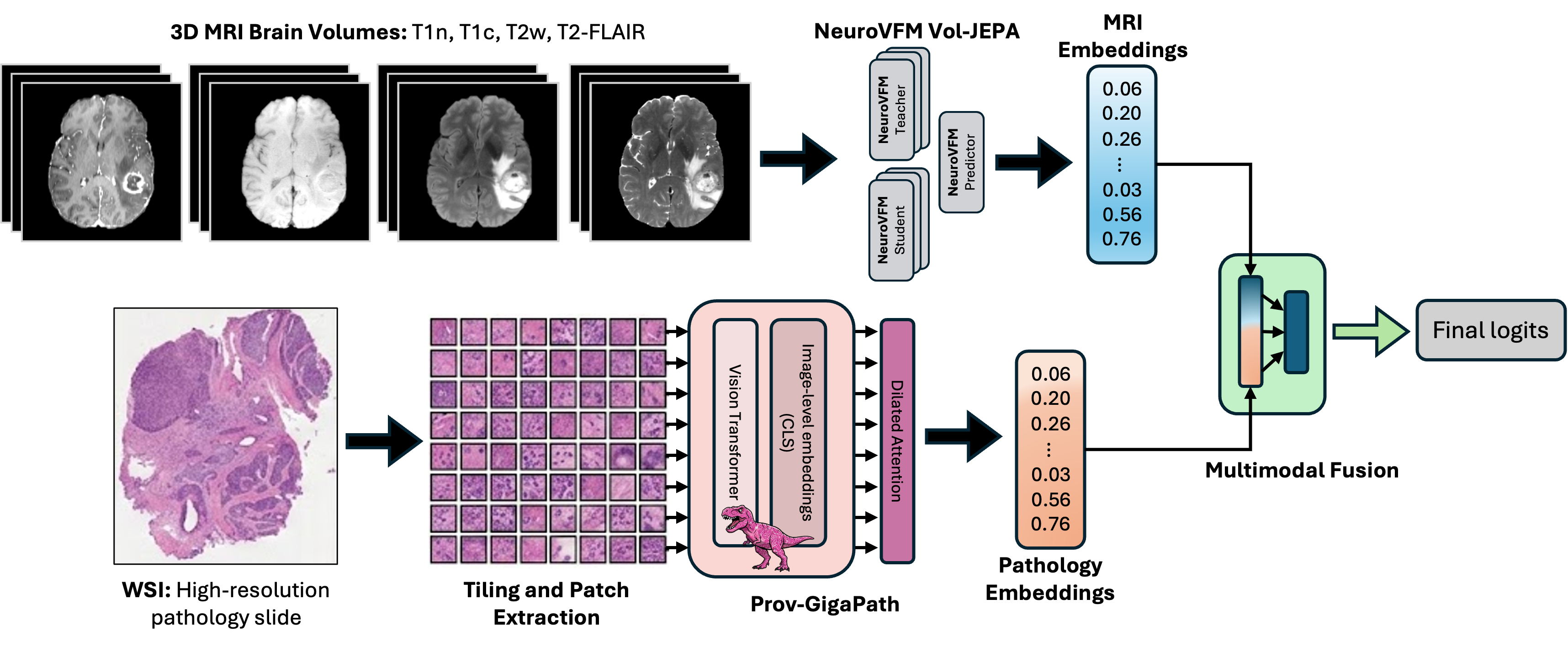}
    \caption{Overview of the proposed CoRe-BT-Fusion framework. Pretrained histopathology and MRI foundation models generate modality-specific embeddings, which are fused for supervised tumor type prediction under variable modality availability.}
    \label{fig:placeholder}
\vspace{-0.6cm}
\end{figure}

\subsection{MRI embedding}

To extract embeddings from neuroimaging, we use NeuroVFM \cite{neurovfm_kondepudi2025health}, a volumetric neuroimaging foundation model trained on 5.24 million MRI and CT 
volumes acquired during routine clinical care. It employs a joint-embedding predictive 
architecture (JEPA) operating directly on 3D volumes, enabling scalable self-supervised learning 
without curated labels or handcrafted preprocessing. In \textit{CoRe-BT-Fusion}, all MRI volumes from a subject are embedded using NeuroVFM, and the subject-level MRI embedding is computed as the mean of the sequence-level embeddings, which is then used as input for downstream classification tasks.

\subsection{Histopathology embedding}
% Add CLAM, HSV color segmentation to eliminate artifacts, mean of the last layer embeddings for multi-slide subjects 
To extract embeddings from whole-slide histopathology imaging, we use Prov-GigaPath \cite{xu2024gigapath}, a whole-slide histopathology foundation model pretrained on 1.3 billion 
$256 \times 256$ tiles extracted from 171,189 whole-slide images spanning over 30,000 patients 
and 31 tissue types. To produce slide embeddings, Prov-Gigapath first requires tiling gigapixel WSI slides into 256x256 tiles. Then, these tiles are embedded using a fine-tuned DINO-v2 encoder. Finally, a LongNet-adapted vision transformer, is applied for ultra-long context modeling over the tile embeddings, allowing for attention over tens of thousands of tiles. This process produces whole slide embeddings that preserve both 
local cellular morphology and global tissue architecture. In the following experiments, for subjects with multiple slides, the subject-level embedding was computed as the mean of the final-layer slide embeddings.

\subsection{Multimodal feature fusion}
For multimodal feature fusion, we first train linear probes for each foundation model on their respective modality, for each task. Then, the fusion module combines MRI and histopathology embeddings using instance-specific weights $w_i$ which indicate how much each modality will contribute to the final prediction. For additional learning, a learnable scalar gate, $\sigma(\alpha)$ is used to weight the contribution of the residual branch. By zero-initializing the residual branch, $\Phi$, the model initially behaves like an ensemble of modality experts, encouraging the final prediction to improve on this baseline. The final prediction, $y$ can be expressed as:

\begin{equation}
y = \sum_{i \in \{MRI, Histo\}} 
\left( w_i \cdot \text{Probe}_i(\mathbf{e}_i) \right)
+ \sigma(\alpha) \cdot 
\Phi\big([\mathbf{e}_{MRI} \cdot m_{MRI}; 
\mathbf{e}_{Histo} \cdot m_{Histo}]\big).
\end{equation}.

% \begin{table}[H]
% \centering
% \caption{
% Subject-level multiclass classification performance using image-derived features.
% Models are trained on 184 subjects and evaluated on a held-out test set of 45 subjects.
% Overall accuracy is reported on the test set.
% Macro-averaged metrics (accuracy, precision, recall, and F1) are computed as the
% unweighted mean across all four WHO grades.
% Best values per column are highlighted in \textcolor{blue}{blue}.
% }
% \label{tab:corebt_feature_classification}

% \resizebox{\linewidth}{!}{
% \begin{tabular}{lccccc}
% \toprule
% \textbf{Model} &
% \textbf{Accuracy} &
% \textbf{Macro Acc.} &
% \textbf{Precision (Macro)} &
% \textbf{Recall (Macro)} &
% \textbf{F1 (Macro)} \\
% \midrule
% Logistic Regression (L2) & 0.578 & \textcolor{blue}{0.455} & 0.414 & \textcolor{blue}{0.455} & \textcolor{blue}{0.429} \\
% Linear SVM               & 0.578 & 0.421 & \textcolor{blue}{0.442} & 0.421 & 0.428 \\
% RBF SVM                  & 0.578 & 0.326 & 0.289 & 0.326 & 0.305 \\
% Random Forest            & 0.644 & 0.242 & 0.173 & 0.242 & 0.201 \\
% Gradient Boosting        & 0.644 & 0.269 & 0.225 & 0.269 & 0.242 \\
% KNN                      & 0.689 & 0.313 & 0.348 & 0.313 & 0.307 \\
% Linear Probe                      & \textcolor{blue}{0.696} & -- & -- & -- & -- \\
% \bottomrule
% \end{tabular}
% }
% \end{table}

\section{Evaluation Metrics}

Tumor typing performance was evaluated at the subject level using standard multiclass classification metrics designed to account for class imbalance and heterogeneous tumor subtype prevalence. The following macro metrics were used:

% \begin{itemize}[topsep=0pt, partopsep=0pt, parsep=0pt, itemsep=1pt, leftmargin=1.2em]

% \vspace{-4pt}
% \[
% \mathrm{Accuracy_{macro}} =
% \frac{1}{C} \sum_{c=1}^{C}
% \frac{\mathrm{TP}_c}{\mathrm{TP}_c + \mathrm{FN}_c},
% \quad
% \mathrm{Precision}_{\mathrm{macro}} =
% \frac{1}{C} \sum_{c=1}^{C}
% \frac{\mathrm{TP}_c}{\mathrm{TP}_c + \mathrm{FP}_c},
% \]
% \[
% \mathrm{Recall}_{\mathrm{macro}} =
% \frac{1}{C} \sum_{c=1}^{C}
% \frac{\mathrm{TP}_c}{\mathrm{TP}_c + \mathrm{FN}_c},
% \quad
% \mathrm{F1}_{\mathrm{macro}} =
% \frac{1}{C} \sum_{c=1}^{C}
% \frac{2\,\mathrm{TP}_c}
% {2\,\mathrm{TP}_c + \mathrm{FP}_c + \mathrm{FN}_c}
% \]
% \vspace{-8pt}

% \end{itemize}

$\text{Accuracy} = \frac{1}{C}\sum (\text{TP}_c + \text{TN}_c)/\text{T}_c$, 
$\text{Precision} = \frac{1}{C}\sum \text{TP}_c/(\text{TP}_c + \text{FP}_c)$, 
$\text{Recall} = \frac{1}{C}\sum \text{TP}_c/(\text{TP}_c + \text{FN}_c)$, and 
$\text{F1} = \frac{1}{C}\sum 2\text{TP}_c/(2\text{TP}_c + \text{FP}_c + \text{FN}_c)$. 
Here, $\mathrm{TP}_c$, $\mathrm{FP}_c$, and $\mathrm{FN}_c$ denote the number of true positives, false positives, and false negatives for class $c$, respectively; $N$ is the total number of subjects and $C$ is the number of tumor classes.

\section{Experiments and Results}

We evaluate the \textit{CoRe-BT-Fusion} on three tumor typing tasks: 1) WHO Grade prediction (3-class), 2) binary low-grade versus high-grade glioma (LGG vs HGG), and 3) Level-1 tumor grouping reflecting biologically meaningful molecular subtypes (4 class). As baseline, we compare modality-specific expert linear probes from the foundation model embeddings to our proposed fusion approach. The test subject cohort consists of the same subjects across tasks, and all subjects contain both modalities to support the modality ablation study. 

\label{subsec:results}

\begin{table}[H]
\centering
\caption{Modality ablation study for LGG-HGG classification.}
\label{tab:lgg_hgg_split}
\renewcommand{\arraystretch}{1.2}
\resizebox{0.9\linewidth}{!}{
\begin{tabular}{|c|l|c|c|c|c|}
\hline
\textbf{Task} & \textbf{Model Configuration} & \textbf{Acc. (Macro)} & \textbf{Prec. (Macro)} & \textbf{Rec. (Macro)} & \textbf{F1 (Macro)} \\ \hline
\multirow{6}{*}{\rotatebox{90}{\textbf{LGG-HGG}}} 
& MRI only & 0.812 & 0.727 & 0.812 & 0.757 \\ \cline{2-6} 
& Pathology only & 0.812 & 0.727 & 0.812 & 0.757 \\ \cline{2-6} 
& Fusion & 0.792 & 0.690 & 0.792 & 0.717 \\ \cline{2-6} 
& Fusion (MRI ablated) & 0.792 & 0.690 & 0.792 & 0.717 \\ \cline{2-6} 
 & Fusion (Pathology ablated) & \textbf{0.833} & \textbf{0.778} & \textbf{0.833} & \textbf{0.801} \\ \cline{2-6} 
& Fusion (Both ablated) & 0.812 & 0.727 & 0.812 & 0.757 \\ \hline
\end{tabular}
}
\end{table}

\begin{table}[H]
\centering
\caption{Modality ablation study for Level 1 Class and WHO Grade classification.}
\label{tab:multi_task_split}
\renewcommand{\arraystretch}{1.2}
\resizebox{0.9\linewidth}{!}{
\begin{tabular}{|c|l|c|c|c|c|}
\hline
\textbf{Task} & \textbf{Model Configuration} & \textbf{Acc. (Macro)} & \textbf{Prec. (Macro)} & \textbf{Rec. (Macro)} & \textbf{F1 (Macro)} \\ \hline

\multirow{6}{*}{\rotatebox{90}{\textbf{Level 1 Class}}} 
& MRI only & 0.470 & 0.488 & 0.470 & 0.467 \\ \cline{2-6} 
& Pathology only & 0.470 & 0.488 & 0.470 & 0.467 \\ \cline{2-6} 
 & Fusion & \textbf{0.556} & \textbf{0.595} & \textbf{0.556} & \textbf{0.560} \\ \cline{2-6} 
& Fusion (MRI ablated) & 0.525 & 0.523 & 0.525 & 0.520 \\ \cline{2-6} 
& Fusion (Pathology ablated) & 0.269 & 0.292 & 0.269 & 0.255 \\ \cline{2-6} 
& Fusion (Both ablated) & 0.470 & 0.488 & 0.470 & 0.467 \\ \hline

\multicolumn{6}{c}{\vspace{-0.2cm}} \\ \hline 

\multirow{6}{*}{\rotatebox{90}{\textbf{WHO Grade}}} 
& MRI only & 0.567 & 0.621 & 0.567 & 0.579 \\ \cline{2-6} 
& Pathology only & 0.600 & 0.558 & 0.600 & 0.569 \\ \cline{2-6} 
& Fusion & 0.600 & 0.558 & 0.600 & 0.569 \\ \cline{2-6} 
& Fusion (MRI ablated) & 0.600 & 0.558 & 0.600 & 0.569 \\ \cline{2-6} 
& Fusion (Pathology ablated) & \textbf{0.667} & \textbf{0.656} & \textbf{0.667} & \textbf{0.619} \\ \cline{2-6} 
& Fusion (Both ablated) & 0.600 & 0.558 & 0.600 & 0.569 \\ \hline
\end{tabular}
}
\end{table}

\section{Discusssion}
The results presented in Section \ref{subsec:results} indicate that multimodal training using the \textit{CoRe-BT-Fusion} approach improves macro accuracy, precision, recall, and F1-score over the modality expert linear probe approaches. Interestingly, for LGG/HGG and WHO Grade classification, the \textit{CoRe-BT-Fusion} model with pathology ablated surpassed the fusion model. This indicates that the model leverages insights gained from multimodal training to improve interpretation of histopathology imaging. Finally, for the Level 1 clinical classification, the multimodal input \textit{CoRe-BT-Fusion} model outperformed all variants. These results demonstrate that multimodal integration offers the greatest benefit for granular tumor subtype classification, where complementary information across modalities improves predictive performance.

%- multimodal training improved accuracy, precision, recall and f1 for all of the proposed tests. 
%- For LGG/HGG and WHO the fusion trained model with pathology ablated surpassed the fusion model.
%- For Level 1 classification the fusion model surpassed the rest.
%- This indicates that for more granular clinical classifications, the multimodal model produces the most reliable predictions. 

\section{Conclusion}
%We present CoReBT Data, a 

We introduced CoRe-BT, a clinically grounded multimodal benchmark for fine-grained glioma tumor typing that integrates MRI, whole-slide histopathology, and diagnostic text under variable modality availability. By reflecting realistic neuro-oncologic workflows, the benchmark enables systematic evaluation of cross-modal learning in incomplete-data settings. Baseline results demonstrate the promise of multimodal fusion and highlight the complementary value of radiology and pathology representations. We hope that CoRe-BT will foster continued research toward robust, clinically meaningful multimodal AI systems for brain tumor diagnosis.

% \begin{table}
% \caption{Table captions should be placed above the
% tables.}\label{tab1}
% \begin{tabular}{|l|l|l|}
% \hline
% Heading level &  Example & Font size and style\\
% \hline
% Title (centered) &  {\Large\bfseries Lecture Notes} & 14 point, bold\\
% 1st-level heading &  {\large\bfseries 1 Introduction} & 12 point, bold\\
% 2nd-level heading & {\bfseries 2.1 Printing Area} & 10 point, bold\\
% 3rd-level heading & {\bfseries Run-in Heading in Bold.} Text follows & 10 point, bold\\
% 4th-level heading & {\itshape Lowest Level Heading.} Text follows & 10 point, italic\\
% \hline
% \end{tabular}
% \end{table}

\bibliographystyle{splncs04}
\bibliography{bibliography}

@article{xu2024gigapath,
  title={{A whole-slide foundation model for digital pathology from real-world data}},
  author={Xu, Hanwen and Usuyama, Naoto and Bagga, Jaspreet and Zhang, Sheng and Rao, Rajesh and Naumann, Tristan and Wong, Cliff and Gero, Zelalem and González, Javier and Gu, Yu and Xu, Yanbo and Wei, Mu and Wang, Wenhui and Ma, Shuming and Wei, Furu and Yang, Jianwei and Li, Chunyuan and Gao, Jianfeng and Rosemon, Jaylen and Bower, Tucker and Lee, Soohee and Weerasinghe, Roshanthi and Wright, Bill J. and Robicsek, Ari and Piening, Brian and Bifulco, Carlo and Wang, Sheng and Poon, Hoifung},
  journal={Nature},
  year={2024},
  publisher={Nature Publishing Group UK London}
}

@misc{chen2022hipt,
      title={{Scaling Vision Transformers to Gigapixel Images via Hierarchical Self-Supervised Learning}}, 
      author={Richard J. Chen and Chengkuan Chen and Yicong Li and Tiffany Y. Chen and Andrew D. Trister and Rahul G. Krishnan and Faisal Mahmood},
      year={2022},
      eprint={2206.02647},
      archivePrefix={arXiv},
      primaryClass={cs.CV},
      url={https://arxiv.org/abs/2206.02647}, 
}

@article{Louis-2021,
    author = {Louis, David N and Perry, Arie and Wesseling, Pieter and Brat, Daniel J and Cree, Ian A and Figarella-Branger, Dominique and Hawkins, Cynthia and Ng, H K and Pfister, Stefan M and Reifenberger, Guido and Soffietti, Riccardo and von Deimling, Andreas and Ellison, David W},
    title = {{The 2021 WHO Classification of Tumors of the Central Nervous System: a summary}},
    journal = {Neuro-Oncology},
    volume = {23},
    number = {8},
    pages = {1231-1251},
    year = {2021},
    month = {06},
    issn = {1522-8517},
    doi = {10.1093/neuonc/noab106},
    url = {https://doi.org/10.1093/neuonc/noab106},
    eprint = {https://academic.oup.com/neuro-oncology/article-pdf/23/8/1231/39535372/noab106.pdf},
}

@article{Price-2024,
    author = {Price, Mackenzie and Ballard, Christine and Benedetti, Julia and Neff, Corey and Cioffi, Gino and Waite, Kristin A and Kruchko, Carol and Barnholtz-Sloan, Jill S and Ostrom, Quinn T},
    title = {{CBTRUS Statistical Report: Primary Brain and Other Central Nervous System Tumors Diagnosed in the United States in 2017–2021}},
    journal = {Neuro-Oncology},
    volume = {26},
    number = {Supplement 6},
    pages = {vi1-vi85},
    year = {2024},
    month = {10},
    issn = {1522-8517},
    doi = {10.1093/neuonc/noae145},
    url = {https://doi.org/10.1093/neuonc/noae145},
    eprint = {https://academic.oup.com/neuro-oncology/article-pdf/26/Supplement_6/vi1/59464369/noae145.pdf}
}

@article{medimageinsight2024,
  author       = {Noel C. F. Codella and
                  Ying Jin and
                  Shrey Jain and
                  Yu Gu and
                  Ho Hin Lee and
                  Asma {Ben Abacha} and
                  Alberto Santamar{\'{\i}}a{-}Pang and
                  Will Guyman and
                  Naiteek Sangani and
                  Sheng Zhang and
                  Hoifung Poon and
                  Stephanie L. Hyland and
                  Shruthi Bannur and
                  Javier Alvarez{-}Valle and
                  Xue Li and
                  John W. Garrett and
                  Alan B. McMillan and
                  Gaurav Rajguru and
                  Madhu Maddi and
                  Nilesh Vijayrania and
                  Rehaan Bhimai and
                  Nick Mecklenburg and
                  Rupal Jain and
                  Daniel Holstein and
                  Naveen Gaur and
                  Vijay Aski and
                  Jenq{-}Neng Hwang and
                  Thomas Lin and
                  Ivan Tarapov and
                  Matthew P. Lungren and
                  Mu Wei},
  title        = {{MedImageInsight: An Open-Source Embedding Model for General Domain
                  Medical Imaging}},
  journal      = {CoRR},
  volume       = {abs/2410.06542},
  year         = {2024},
  url          = {https://doi.org/10.48550/arXiv.2410.06542}
}

@article{yuan2021florence,
  author       = {Lu Yuan and
                  Dongdong Chen and
                  Yi{-}Ling Chen and
                  Noel Codella and
                  Xiyang Dai and
                  Jianfeng Gao and
                  Houdong Hu and
                  Xuedong Huang and
                  Boxin Li and
                  Chunyuan Li and
                  Ce Liu and
                  Mengchen Liu and
                  Zicheng Liu and
                  Yumao Lu and
                  Yu Shi and
                  Lijuan Wang and
                  Jianfeng Wang and
                  Bin Xiao and
                  Zhen Xiao and
                  Jianwei Yang and
                  Michael Zeng and
                  Luowei Zhou and
                  Pengchuan Zhang},
  title        = {{Florence: A New Foundation Model for Computer Vision}},
  journal      = {CoRR},
  volume       = {abs/2111.11432},
  year         = {2021},
  url          = {https://arxiv.org/abs/2111.11432}
}

@misc{zhang2025biomedclip,
      title={{BiomedCLIP: a multimodal biomedical foundation model pretrained from fifteen million scientific image-text pairs}}, 
      author={Sheng Zhang and Yanbo Xu and Naoto Usuyama and Hanwen Xu and Jaspreet Bagga and Robert Tinn and Sam Preston and Rajesh Rao and Mu Wei and Naveen Valluri and Cliff Wong and Andrea Tupini and Yu Wang and Matt Mazzola and Swadheen Shukla and Lars Liden and Jianfeng Gao and Angela Crabtree and Brian Piening and Carlo Bifulco and Matthew P. Lungren and Tristan Naumann and Sheng Wang and Hoifung Poon},
      year={2025},
      eprint={2303.00915},
      archivePrefix={arXiv},
      primaryClass={cs.CV},
      url={https://arxiv.org/abs/2303.00915}, 
}

@inproceedings{RadfordKHRGASAM21,
  author       = {Alec Radford and
                  Jong Wook Kim and
                  Chris Hallacy and
                  Aditya Ramesh and
                  Gabriel Goh and
                  Sandhini Agarwal and
                  Girish Sastry and
                  Amanda Askell and
                  Pamela Mishkin and
                  Jack Clark and
                  Gretchen Krueger and
                  Ilya Sutskever},
  editor       = {Marina Meila and
                  Tong Zhang},
  title        = {{Learning Transferable Visual Models From Natural Language Supervision}},
  booktitle    = {Proceedings of the 38th International Conference on Machine Learning,
                  {ICML} 2021, 18-24 July 2021, Virtual Event},
  series       = {Proceedings of Machine Learning Research},
  volume       = {139},
  pages        = {8748--8763},
  publisher    = {{PMLR}},
  year         = {2021},
  url          = {http://proceedings.mlr.press/v139/radford21a.html} 
}

@inproceedings{Farahani2020CPMRP_miccai20challenge,
  author    = {Farahani, Keyvan and
               Kurc, Tahsin and
               Bakas, Spyridon and
               Bearce, Benjamin Aaron and
               Kalpathy-Cramer, Jayashree and
               Freymann, John and
               Saltz, Joel and
               Stahlberg, Eric and
               Zaki, George and
               Nasrallah, MacLean P. and
               Shinohara, Russell Taki},
  title     = {{Computational Precision Medicine Radiology-Pathology Challenge on Brain Tumor Classification 2020}},
  booktitle = {Proceedings of the 23rd International Conference on Medical Image Computing and Computer Assisted Intervention (MICCAI 2020)},
  year      = {2020},
  address   = {Lima, Peru},
  publisher = {Zenodo},
  doi       = {10.5281/zenodo.3718894},
  url       = {https://doi.org/10.5281/zenodo.3718894}
}

@article{ferreira2024we,
  title={{How we won BraTS 2023 adult glioma challenge? Just faking it! enhanced synthetic data augmentation and model ensemble for brain tumour segmentation}},
  author={Ferreira, Andr{\'e} and Solak, Naida and Li, Jianning and Dammann, Philipp and Kleesiek, Jens and Alves, Victor and Egger, Jan},
  journal={arXiv preprint arXiv:2402.17317},
  year={2024}
}

@article{baid2021rsna,
  title={{The RSNA-ASNR-MICCAI BraTS 2021 benchmark on brain tumor segmentation and radiogenomic classification}},
  author={Baid, Ujjwal and Ghodasara, Satyam and Mohan, Suyash and Bilello, Michel and Calabrese, Evan and Colak, Errol and Farahani, Keyvan and Kalpathy-Cramer, Jayashree and Kitamura, Felipe C and Pati, Sarthak and others},
  journal={arXiv preprint arXiv:2107.02314},
  year={2021}
}

@article{Davatzikos2018CIPT,
  author       = {C. Davatzikos and S. Rathore and S. Bakas and S. Pati and M. Bergman and R. Kalarot and P. Sridharan and A. Gastounioti and N. Jahani and E. Cohen and H. Akbari and B. Tunc and J. Doshi and D. Parker and M. Hsieh and A. Sotiras and H. Li and Y. Ou and R. K. Doot and M. Bilello and Y. Fan and R. T. Shinohara and P. Yushkevich and R. Verma and D. Kontos},
  title        = {{Cancer imaging phenomics toolkit: quantitative imaging analytics for precision diagnostics and predictive modeling of clinical outcome}},
  journal      = {Journal of Medical Imaging},
  volume       = {5},
  number       = {1},
  pages        = {011018},
  year         = {2018},
  doi          = {10.1117/1.JMI.5.1.011018}
}

@inproceedings{Pati2020CaPTk,
  author       = {S. Pati and A. Singh and S. Rathore and A. Gastounioti and M. Bergman and P. Ngo and S. M. Ha and D. Bounias and J. Minock and G. Murphy and H. Li and A. Bhattarai and A. Wolf and P. Sridaran and R. Kalarot and H. Akbari and A. Sotiras and S. P. Thakur and R. Verma and R. T. Shinohara and P. Yushkevich and Y. Fan and D. Kontos and C. Davatzikos and S. Bakas},
  title        = {{The Cancer Imaging Phenomics Toolkit (CaPTk): Technical Overview}},
  booktitle    = {BrainLes 2019. Lecture Notes in Computer Science},
  volume       = {11993},
  pages        = {380--394},
  publisher    = {Springer},
  year         = {2020},
  doi          = {10.1007/978-3-030-46643-5\_38}
}

@article{isensee2021nnunet,
  title   = {{nnU-Net: A Self-Configuring Method for Deep Learning-Based Biomedical Image Segmentation}},
  author  = {Isensee, Fabian and Jaeger, Paul F. and Kohl, Simon A. A. and Petersen, Jens and Maier-Hein, Klaus H.},
  journal = {Nature Methods},
  volume  = {18},
  number  = {2},
  pages   = {203--211},
  year    = {2021},
  doi     = {10.1038/s41592-020-01008-z}
}

@article{CLAMlu2021data,
  title={Data-efficient and weakly supervised computational pathology on whole-slide images},
  author={Lu, Ming Y and Williamson, Drew FK and Chen, Tiffany Y and Chen, Richard J and Barbieri, Matteo and Mahmood, Faisal},
  journal={Nature Biomedical Engineering},
  volume={5},
  number={6},
  pages={555--570},
  year={2021},
  publisher={Nature Publishing Group}
}

@article{neurovfm_kondepudi2025health,
  title={Health system learning achieves generalist neuroimaging models},
  author={Kondepudi, Akhil and Rao, Akshay and Zhao, Chenhui and Lyu, Yiwei and Harake, Samir and Banerjee, Soumyanil and Joshi, Rushikesh and Meissner, Anna-Katharina and Hou, Renly and Jiang, Cheng and others},
  journal={arXiv preprint arXiv:2511.18640},
  year={2025}
}

\end{document}